\documentclass[runningheads]{llncs}
\usepackage{subcaption}
\usepackage{algorithm}
\usepackage{algorithmic}
\usepackage{amsmath}
\usepackage{multirow}
\usepackage{amssymb}
\usepackage{amsfonts}
\usepackage{booktabs}
\usepackage{hyperref}
\usepackage{xcolor} 
\usepackage{siunitx}
\newcommand{\capX}{\mathbf{X}}
\newcommand{\capT}{\mathbf{T}}

\usepackage[T1]{fontenc}
\usepackage{graphicx,verbatim}

\begin{document}
\title{3D Densification for Multi-Map Monocular VSLAM in Endoscopy }

\author{X. Anadón, Javier Rodríguez-Puigvert, J.M.M. Montiel} 
\institute{Universidad de Zaragoza, Zaragoza, Spain \\
    \email{\{xanadon,jrp,josemari\}@unizar.es}\\ 
    \textit{Preprint}}

\maketitle        

\begin{abstract}
Multi-map Sparse Monocular visual Simultaneous Localization and Mapping applied to monocular endoscopic sequences has proven efficient to robustly recover tracking after the frequent losses in endoscopy due to motion blur, temporal occlusion, tools interaction or water jets. The sparse multi-maps are adequate for robust camera localization, however, they are very poor for environment representation. They are noisy, with a high percentage of inaccurately reconstructed 3D points, including significant outliers, and more importantly, with an unacceptable low density for clinical applications. 

We propose a method to remove outliers and densify the maps of the state of the art for sparse endoscopy multi-map CudaSIFT-SLAM. The NN LightDepth for up-to-scale depth dense predictions are aligned with the sparse CudaSIFT submaps by means of the robust to spurious LMedS.
Our system mitigates the inherent scale ambiguity in monocular depth estimation while filtering outliers, leading to reliable densified 3D maps. 

We provide experimental evidence of accurate densified maps \qty{4.15}{\milli\metre} RMS accuracy at affordable computing time in the C3VD phantom colon dataset. We report qualitative results on the real colonoscopy from the Endomapper dataset.

\end{abstract}
\section{Introduction}

Dense 3D reconstruction of endoscopic images is critical for enhancing visualization in medical procedures, assisting clinicians during interventions, and enabling autonomous robotic systems. 3D dense representation of the organ could help physicians assess lesion sizes, ensure complete mucosa coverage, and minimize the risk of missing critical areas during exploration \cite{olympusAR}.
However, the endoscopy environment is challenging, characterized by low-texture regions, dynamic lighting, and abrupt camera movements, making it particularly challenging to obtain reliable 3D dense reconstructions. 

Multi-maps feature-based Monocular visual Simultaneous Localization and Mapping achieve accurate camera tracking, producing sparse 3D reconstructions that are too noisy and lacking in detail for clinical use. CudaSIFT-SLAM\cite{elvira2024} has demonstrated real-time performance and robust camera tracking, nevertheless, their sparse representation limits its applicability.
To overcome these limitations, we propose a 3D dense multi-mapping system that combines the robust camera tracking of a feature-based SLAM with the dense depth prediction of a single-view self-supervised depth estimation. 
Our system integrates the sparse maps from CudaSIFT-SLAM with the up-to-scale predictions of LightDepth\cite{Rodriguez-Puigvert_2023_ICCV} aligning them using Least Median Squares(LMedS)\cite{LMS} for scale corrections and outlier filtering. Our method achieves a densification of the multi-maps while maintaining the camera pose accuracy during the endoscope trajectory.  

We propose a robust method that aligns dense up-to-scale predictions with sparse points to achieve dense reconstructions. It has the following contributions: 
\begin{itemize} 
    \item Dense 3D surface reconstruction in less than 200 ms, which makes it suitable for real-time applications. 
    
    \item Experimental validation on the C3VD dataset, including full screening sequences, and qualitative tests on real colonoscopy recordings from the Endomapper dataset. 
    
    \item Removing the need for domain adaptation, as the components of the system are agnostic to domain specifications.
\end{itemize}

\section{Related Work}
Feature-based SLAM \cite{murAcceptedTRO2015} has been successfully applied to abdominal endoscopy in \cite{mahmoud2018live} computing reliable 3D poses for the video frames as a first step towards a multiview stereo densification based on correlation matching between pixel frames. It is assumed a rigid scene.
Focused on endoscopic endonasal surgery, SAGE SLAM \cite{liu2022sage} proposes to use a depth network \cite{unet} that corrects a global scale in a deep feature SLAM pipeline.
Discrete features have also been exploited to handle deforming tissue, NR-SLAM \cite{nrslam} combines a Dynamic Deformation Graph with a Visco-Elastic deformation model, allowing automatic initialization and extension of a sparse point cloud in deforming environments.

However, feature-based systems struggled with low-texture regions and feature scarcity, limiting their robustness for real procedures, especially in colonoscopy. In any case, CudaSIFT-SLAM \cite{elvira2024}, based on ORB-SLAM3 \cite{ORBSLAM3_TRO} leveraging CudaSIFT features, has demonstrated real-time performance and high accuracy in recovering the camera pose by incorporating a quasi-rigid deformation model for monocular colonoscopy sequences. 
CudaSIFT-SLAM \cite{elvira2024} ability to merge multiple maps and perform relocalization enables tracking recovery, making it particularly effective for exploratory trajectories in endoscopy, where unpredictable motion and occlusions often cause tracking failures. In clinical settings, feature-based SLAM is restricted due to the sparse representation where a dense representation is essential.

Direct methods have also been explored to achieve 3D dense reconstruction in endoscopy, mainly based on the minimization of a photometric residual for camera tracking in combination with a depth estimation network for densification. Endo-Depth\cite{endodepth} uses \cite{monodepth2} with photometric residual-based camera tracking. RNN-SLAM \cite{ma2021rnnslam} integrates DSO \cite{dso} with a recurrent neural network, leveraging temporal dependencies to refine depth and pose estimation \cite{wang2019recurrent}. Direct methods, however, lack loop closing detection, multi-mapping, and are unable to recover from camera tracking losses, which limits their performance in realistic colonoscopy scenes where tracking losses and map fragmentation are prevalent. 

Depth perception plays a crucial role when it comes to 3D dense reconstructions from endoscopic images. Supervised methods rely on annotated datasets\cite{Visentini-Scarzanella2017}, and the lack of depth-annotated endoscopic data limits their use. Multi-view self-supervision has been addressed for endoscopic images \cite{luo2019details,huang2021self}, but remains challenging due to the presence of deformations, weak texture and specular reflections. Synthetic-to-real domain adaptation has been addressed with GAN models\cite{rau2019implicit,karaoglu2021adversarial,Wang2024StructurePreserving} and with teacher-student architectures\cite{Rodriguez2022,paruchuri2024leveraging}. 
Single-View self-supervised learning has emerged \cite{Rodriguez-Puigvert_2023_ICCV} exploiting the illumination decline as a natural depth cue in the endoscopic setting where the camera and light are co-located. In contrast to multi-view methods and supervised approaches, LightDepth can be trained directly from individual frames, providing a scalable solution without the need for explicit ground truth depth annotations.

The usage of robust estimators to align the scale of dense depth estimations to the sparse 3D point resulting from SfM can be traced back to \cite{luo2020consistent}, in which the median depth is proposed,  \cite{izquierdo2023sfm} goes one step further using RANSAC to compute a robust per keyframe scale correction. Closer to us is \cite{Lalaguna2024tfg} where the LMedS scale alignment is proposed for the first time in the off-line processing of gastroscopies to measure the Barrett esophagus. 

3D Dense Reconstructions based on NeRFs and 3D Gaussian splatting have also been explored for endoscopic images, although with a primary focus on rendering quality, rather than explicitly modeling the geometric accuracy of the reconstructed scene. 

Wang et al.\cite{wang2022neural} propose the use of NeRFs for known cameras and stereo depth in deformable laparoscopic scenes. FreeSurgGS\cite{Guo2024FreeSurGS} proposes a combination of SfM and 3D Gaussian splatting for optimization of the camera poses and scene representation. EndoGSLAM\cite{wang2024endogslam} proposes a dense RGB-D SLAM approach based on the 3D Gaussian representation and differentiable rasterization for tracking and tissue reconstruction.
However, these methods rely on stereo or depth sensors, which are not available in standard monocular endoscopes.
In contrast, we propose a monocular-based approach tailored for conventional endoscopic imaging without additional depth sensing hardware.
Beltran et al. \cite{NFL-BA} introduced bundle adjustment improvements using supervised depth from \cite{paruchuri2024leveraging} into the EndoGSLAM pipeline. Building upon RNN-SLAM\cite{ma2021rnnslam}, Gaussian Pancakes \cite{Bonilla2024ARXIV} improves the 3D Gaussian Splatting representation with geometric generalization. 
Focusing on the quality of 3D reconstruction, LightNeus \cite{batlle2023lightneus} models surfaces with known camera poses. Exploiting the illumination decline principle, it employs a neural implicit representation to compute the 3D surface reconstruction. 

\section{System Overview}
\begin{figure}
    \centering
    \includegraphics[width=\textwidth]{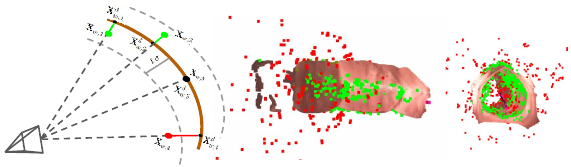}
    \caption{3D dense submap. Left, 2D simplified plot of the LMedS inlier-outlier boundary. Middle and right, two views of the dense model along with the sparse map points.  Green: LMedS inlier sparse map points  after. Red: outliers.}
    \label{fig:method}
\end{figure}

Our system builds a 3D dense multi-map representation of the environment and simultaneously localizes the camera pose with respect to the multi-maps. We built on two systems able to operate in real colonoscopy environments, CudaSIFT-SLAM \cite{elvira2024}, a feature-based monocular SLAM, and LightDepth \cite{Rodriguez-Puigvert_2023_ICCV} a self-supervised single-view depth prediction network based on illumination decline.

CudaSIFT-SLAM multiple-map approach is able to perform with numerous merges that avoid over-segmentation of the sub-maps. Each of those submaps, \(\mathcal{M}\), is defined as a sparse set of 3D points, represented in world reference coordinates as  $\left\{\capX_{w,j} \right\} \subset  \mathbb{R}^3$, and a set of selected frames, $\left\{K_{i}\right\}$, called keyframes, whose localization is defined as a transformation in the world reference $\left\{\capT_{iw}\right\} \subset  \mathrm{SE(3)}$. Each $\mathcal{M}$ has its own scale, as scale is typically not observable in monocular SLAM. These maps can be understood as a summary of visual and geometric information of what was seen during the procedure. Although the submaps $\left\{\capX_{w,j} \right\}$ yielded by CudaSIFT-SLAM are useful for camera pose estimation, they are inherently sparse and noisy. 

For each keyframe $K_{i}$ in a CudaSIFT submap $\mathcal{M}$, LightDepth predicts a depth map $d_i \in (0, \infty)^{w \times h}$ and an albedo map $\rho_i \in [0, 1]^{w \times h \times 2}$.

In the photometric model of LightDepth, the gain is fixed at 1 for all input images, allowing the network to adjust depth estimates $d_i$ to account for variations in brightness, particularly in darker regions resulting in up-to-scale depth estimation with a different scale per each keyframe. To achieve clinically meaningful dense reconstructions, we combine the up-to-scale LightDepth depth maps $d_i$ with the sparse submap points $\left\{\capX_{w,j} \right\}$. 

Per each keyframe \({K_i}\), we compute the scale $s_i$ that aligns its depth map $d_i$ with the 3D sparse point cloud $\left\{\capX_{w,j}\right\}$ minimizing the distance between the sparse map points and the dense depth. The sparse map points are very noisy, and a high fraction of them are spurious. We opt for LMedS because it can compute from scratch the distance threshold to determine if a map point is spurious. Pre-defining a fixed distance threshold is not possible because of the unknown scale for the sparse map, and because of the strong dependence of the 3D point error with the point depth due to the non-linearities of the triangulation, hence RANSAC cannot be applied.

Per each map point $\capX_{w,j}$ we compute a scale proposal:
\begin{equation}
s_{i,j} = \frac{\left\|\capX_{w,j}\right\|}{\left\|\capX_{w,j}^d\right\|}, \;\;\;\;
\capX_{w,j}^d =\pi_i^{-1}(\pi_i(\capT_{iw},\capX_{w,j}),\capT_{iw},d_i)
\end{equation}
Where $\pi_i()$ defines the projection of a map point in camera $i$ and $\pi_i^{-1}()$ yields, $\capX_{w,j}^d$ the unprojection in 3D of an image pixel, assigning the depth according to the depth map $d_i$. 

Each scale proposal is ranked according to the median of the squared residuals for all sparse points in the map, selecting as the scale the one producing the smallest median residual:
\begin{equation}
s_i^{\operatorname{LMedS}}=\arg\min_{s_{i,j}} \operatorname{median}\left(\left\{\epsilon_{i,j,k}^2\right\}\right) k \neq j, \;\;\;
\epsilon_{i,j,k}=\left\|\capX_{w,k}-s_{i,j}\capX_{w,k}^d\right\|
\end{equation}
This first estimate of the scale allows defining the error standard deviation for the inliers as $\sigma = 1.4826 \sqrt{\min_m}$, being $\min_m$ the minimal median squared residual yielded by $s_{i,j}^{\operatorname{LMedS}}$. The $t\sigma$ defines a threshold to identify the inlier map points:
\begin{equation}
\left\|\capX_{w,k}-s_{i,j}^{\operatorname{LMedS}}\capX_{w,k}^d\right\| \geq t \sigma
\end{equation}

The inlier map points are used to refine the scale by means of a non-linear optimization:

\begin{equation}
s_i=\arg\min_{s_{i}} \sum_k \operatorname{r_k}\left(\left\|\capX_{w,k}-s_i\capX_{w,k}^d\right\|^2\right) k\;\text{inlier}\\
\end{equation}

being $\operatorname{r_k}$ a robust influence function and using $s_i^{\operatorname{LMedS}}$ as an initial guess.

We have a scaled RGBD observation per each keyframe. As we also have the pose of the keyframe, we use \textit{Truncated Signed Distance Function (TSDF)} \cite{curless1996volumetric}, in its Open3D implementation \cite{open3d},   to fuse all the scaled RGBD observations in a unique surface. Then the global surface can be explicitly extracted using the \textit{Marching Cubes} algorithm \cite{marching_cubes}.

\begin{figure}
    \centering    
    \includegraphics[width=\linewidth,]{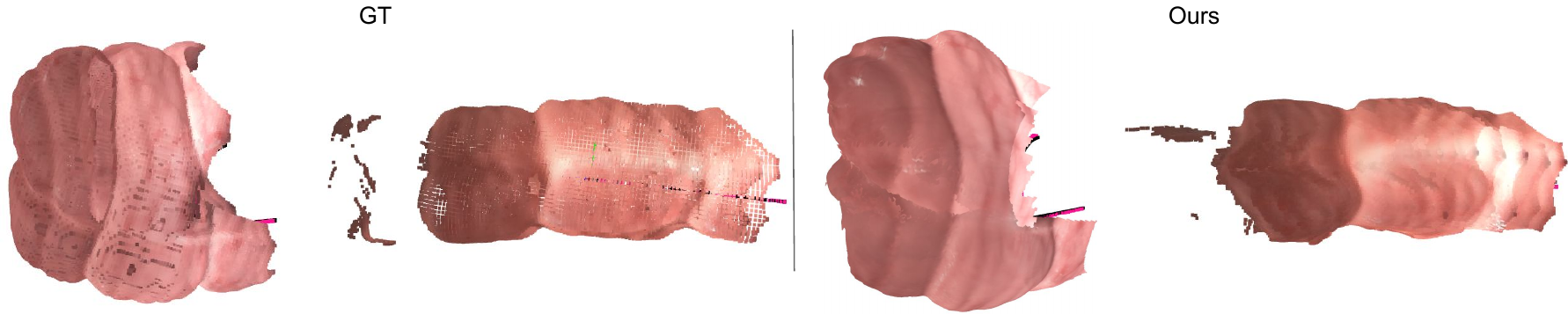}
    \caption{Dense 3D reconstruction for Transverse t2a and Cecum c2a.}
    \label{fig:combined_reconstruction}
\end{figure}
\section{Experimental Results}

\begin{table}[h]
    \centering
    \begin{tabular}{|l|c|c|c|c|c|c|c|c|c|c|c|c|c|c|}
        \hline
        & \multicolumn{9}{c|}{Short Sequences} & \multicolumn{5}{c|}{Screening Sequences} \\
        \cline{2-10} \cline{11-15}
        Metrics & t2a & t3a & t4a & c1a & c2a & c3a & d4a & s3a & Avg & Seq1 & Seq2 & Seq3 & Seq4 & Avg \\
        \cline{1-10}
        \hline
        \# Frames & 194 & 250 & 382 & 276 & 370 & 730 & 148 & 613 & 370 & 5458 & 5100 & 4726 & 4774 & 5015 \\
        \# KF & 16 & 7 & 14 & 21 & 24 & 15 & 11 & 42 & 19 & 467 & 687 & 565 & 381 & 525 \\
        KF time & 243 & 714 & 546 & 263 & 308 & 973 & 269 & 292 & 451 & 234 & 148 & 167 & 251 & 200 \\
        RMS ATE & 0.58 & 0.17 & 0.43 & 0.47 & 0.79 & 0.54 & 0.17 & 0.48 & 0.45 & 3.39 & 2.94 & 2.80 & 3.25 & 3.10 \\
        \hline
    \end{tabular}
    \caption{CudaSIFT-SLAM performance on C3VD. The RMS ATE is in mm and the KF time in ms.}
    \label{tab:CudaSIFTMetrics}
\end{table}

\begin{table}[h]
    \centering
    \begin{tabular}{|c|c|c|c|c|c|c|c|c|c|c|}
        \hline
        {Method} & {{Metrics}} & {t2a} & {t3a} & {t4a} & {c1a} & {c2a} & {c3a} & {d4a} & {s3a} & {avg} \\
        \cline{1-11}
        \hline
        \multirow{3}{*}{CudaSift}& \#MP  ($\times 10^3$) & 1.8 & 1.5 & 2.4 & 1.8 & 2.4. & 2.0. & 2.4 & 1.8 & 2.0 \\
        & RMS Acc. $\downarrow$ & 242.73 & 6.84 & 44.09 & 21.54 & 72.50 & 58.32 & 192.69 & 153.87 & 99.07 \\
        & MedA Acc. $\downarrow$& 2.11 & 1.75 & 1.04 & 3.7 & 1.64 & 3.21 & 1.05 & 1.60 & 2.01 \\
        \hline
        \multirow{3}{*}{Ours}& \#MP ($\times 10^3$) & \num{98} & \num{34} & \num{54} & \num{555} & \num{148} & \num{83} & \num{64} & \num{34} & \num{134} \\
        & RMS Acc. $\downarrow$ & 2.46 & 4.39 & 3.13 & 4.55 & 3.97 & 6.06 & 5.23 & 3.39 & 4.15 \\
        & MedA Acc. $\downarrow$ & 1.68 & 3.15 & 1.70 & 3.13 & 2.42 & 2.75 & 3.70 & 2.26 & 2.60\\
        \hline
        \multirow{2}{*}{LightNeus\cite{batlle2023lightneus}}&RMS Acc. $\downarrow$ & 2.58 & 10.70 & 3.79 & 2.01 & 1.87 & 5.49 & 4.08 & 3.18 & 4.21 \\
        & MedA Acc. $\downarrow$ & 2.24 & 6.39 & 1.15 & 0.95 & 1.40 & 1.12 & 2.66 & 2.57 & 2.31 \\
        \hline
    \end{tabular}
    \caption{Dense maps performance. All the Accuracies: RMS Acc. and MedA Acc. are in mm. }
    \label{tab:resultados-metricas}
\end{table}
\textbf{Quantitative results in C3VD.}
 C3VD balances realism with the availability of accurate ground truth labels. It captures the imaging conditions of a real endoscope, including illumination inverse squared decay, global illumination effects and specular highlights. The dataset includes short sequences and screening sequences. The short sequences contain ground truth in pose, depth and surface normals derived by registering the 2D endoscopic images to corresponding 3D phantom models. The camera trajectory is short, with slow motion, and the camera is far from the colon surfaces.  The screening sequences contain only ground truth camera poses. Regarding the camera movement, it is fast because it mimics the withdrawal phase of a screening colonoscopy. In many cases, there is motion blur and the camera can be very close to the colon surfaces. 

We present the quantitative results on 8 C3VD short test sequences following the train/test split of LightDepth DPT \cite{Rodriguez-Puigvert_2023_ICCV} (See Figure\,\ref{fig:combined_reconstruction}) . We report in Table\,\ref{tab:CudaSIFTMetrics} the performance of CudaSIFT, after alignment with GT by means of the similarity transformation, i.e. rotation, translation and scale that minimizes the ATE between the CudaSIFT estimated camera trajectory and the GT camera trajectory. The camera trajectory achieves a 0.45 mm ATE. On average, the sequences are 370 frames long, are summarized in 19 keyframes, which leaves 375 ms to deal with the densification of each keyframe.

 We  report in Table~\ref{tab:resultados-metricas} the density as the map number of map points (\#MP) either in CudaSIFT and after the proposed densification. Regarding accuracy, we report RMS Accuracy and the MedA Accuracy. The accuracy \cite{Sucar_2021_ICCV} is defined as the distance from each point of our 3D model to its closest in the GT in mm.  The ground truth point cloud is obtained by unprojecting the ground truth depths from ground truth camera poses. For alignment, we use the similarity transformation that aligns the CudaSIFT-SLAM trajectory with the GT trajectory to minimize the ATE.  We also include as baseline LightNeus \cite{batlle2023lightneus} a state-of-the-art 3D reconstruction based on neural rendering exploring the illumination decline principle and coding a dense model with a TSDF as we do.  
 
 Our densification method increases the number of points by a factor bigger than 60. Regarding the accuracy, CudaSIFT-SLAM accuracy is \qty{2.01}{\milli\meter} in MedA but \qty{99}{\milli\meter} in RMS values, showing clearly that the sparse map has a \qty{50}{\percent} of accurate map points, but there are plenty of spurious points that the geometrical methods are not able to remove. When combined with the LightDepth dense depth prediction, both the density and the RMS Accuracy boost. Regarding the comparison with LightNeus, we are able to produce maps with comparable accuracy,at a fraction of the computational cost and not needing the GT camera poses.
Note that, as a baseline, LightDepth reports a per frame RMSE \qty{5.60}{\milli\metre} and MedAE \qty{2.67}{\milli\metre} for single-view depth estimation errors. Thanks to the fusion of accumulative predictions in depth, we achieve a better performance than LightDepth in our after-fusion dense maps.

Regarding the computing time, our densification method requires in total \qty{186}{\milli\second} below the \qty{200}{\milli\second} available per KF. LightDepth inference lasts for \qty{22}{\milli\second}, the LMedS scale alignment \qty{138}{\milli\second} and the TSDF integration \qty{26}{\milli\second}. The Marching Cubes algorithm takes \qty{117}{\milli\second}. All experiments were conducted on an Intel i7 CPU, operating at 3.6GHz, with 32 GB of memory, and utilizing an NVIDIA TITAN V GPU.

\begin{figure}
    \centering    
    \includegraphics[width=0.95\textwidth]{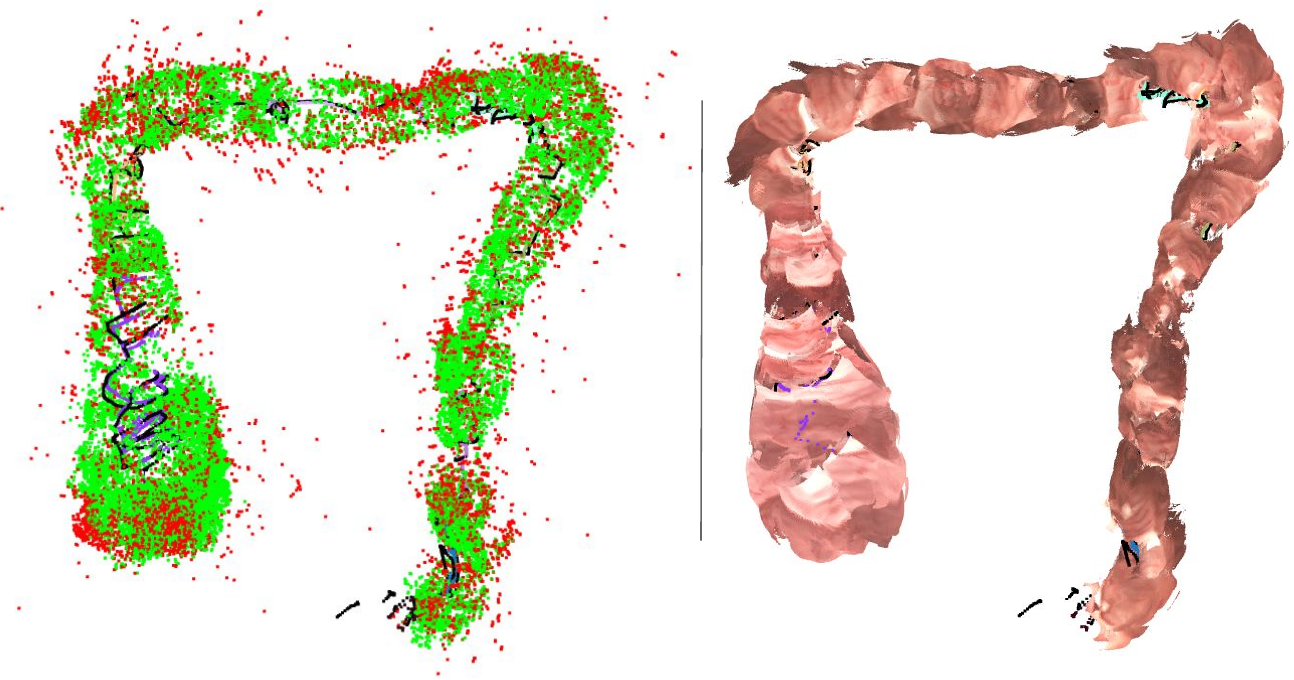}
    \caption{Results for sceening Seq3. Left sparse maps, red outliers, green inliers according to LMedS. Right, dense 3D reconstruction}
    \label{fig:dense-screening}
\end{figure}

When processing the 4 screening sequences, CudaSIFT-SLAM produces multiple maps (19 maps on average) along the trajectory due to camera occlusions and tracking failure. Regarding the trajectory, we reach a \qty{3.09 }{\milli\metre} RMS ATE for the camera trajectory while locating in the map $83.32 \%$ of the frames in each sequence.  Compared to the \qty{0.45 }{\milli\metre} RMS ATE  obtained in short sequences (Table\,\ref{tab:CudaSIFTMetrics})  we observe a significant increase, showing the greater challenge of the screening explorations.  Figure \ref{fig:dense-screening} shows our 3D dense map for Seq3. It is also displayed the capability to detect and reject spurious points in the CudaSIFT-SLAM sparse map.  

\textbf{Qualitative results}
\begin{figure*}
\centering
  \includegraphics[width=\textwidth]{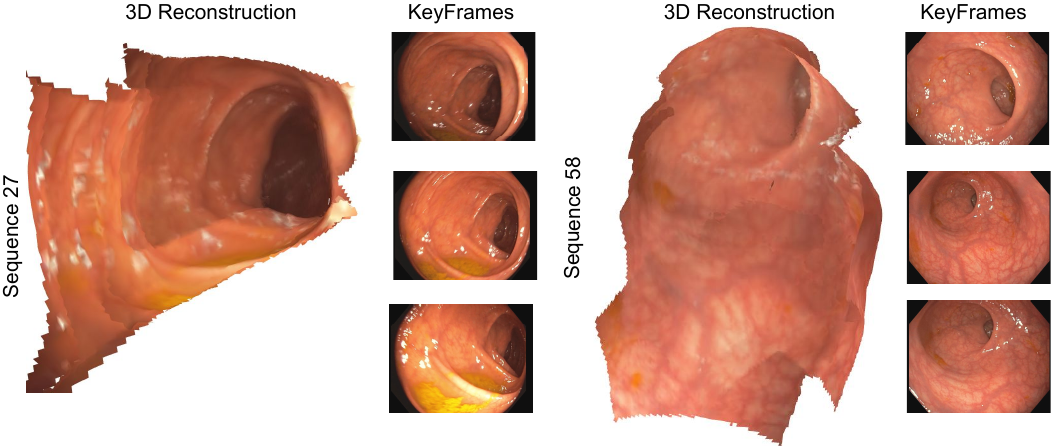}
\caption{Dense maps in EndoMapper in Seq\_027 and Seq\_058}
\label{fig:real}
\end{figure*}
We have conducted qualitative experiments on the EndoMapper dataset ~\cite{azagra2022endomapper}, because it includes real screening procedures with water jets, occlusions, deformation or motion blur. Additionally, an accurate colonoscope calibration is provided, both geometric and photometric. We have trained LightDepth DPT\cite{Rodriguez-Puigvert_2023_ICCV} on \num{42384}  EndoMapper \cite{azagra2022endomapper} images selected using covisibility according to ColonMapper\cite{morlana2024colonmapper}.

We provide qualitative results (Figure\,\ref{fig:real}) in sequences Seq\_027 and Seq\_038 from the EndoMapper dataset, building 3D models from the sole input of the images coming from a standard monocular colonoscope. 

\section{Conclusions}
The proposed CudaSIFT-SLAM + LightDepth combination has proven able to successfully clean from spurious and densify the sparse multi-maps that enable robust camera tracking in real endoscopies. 

Our approach addresses the inherent scale ambiguity in monocular depth estimation NN's by robustly aligning dense up-to-scale depth predictions to the single scale of the sparse 3D map by means of a Least Median Squares strategy followed by a non-linear optimization. Experimental results on the C3VD and Endomapper datasets demonstrate that our method achieves competitive accuracy against state-of-the-art techniques. 

Future work we will explore the integration of the NN depth estimates into the sequential real-time V-SLAM pipeline, including the initialization, relocation and bundle adjustment for the local mapping.

\newpage
\bibliographystyle{splncs04}
\bibliography{biblio}

@article{marching_cubes,
author = {Lorensen, William E. and Cline, Harvey E.},
title = {Marching cubes: A high resolution 3D surface construction algorithm},
year = {1987},
issue_date = {July 1987},
publisher = {Association for Computing Machinery},
address = {New York, NY, USA},
volume = {21},
number = {4},
issn = {0097-8930},
journal = {SIGGRAPH Comput. Graph.},
month = aug,
pages = {163–169},
numpages = {7}
}

@article{ma2021rnnslam,
  title={{RNNSLAM}: Reconstructing the {3D} colon to visualize missing regions during a colonoscopy},
  author={Ma, Ruibin and Wang, Rui and Zhang, Yubo and Pizer, Stephen and McGill, Sarah K and Rosenman, Julian and Frahm, Jan-Michael},
  journal={Medical image analysis},
  volume={72},
  pages={102100},
  year={2021},
  publisher={Elsevier}
}

@article{azagra2022endomapper,
  author = {Azagra, Pablo and et al.},  
  title = {{EndoMapper} dataset of complete calibrated endoscopy procedures},
  journal={Scientific Data},
  volume={10},
  number={1},
  pages={671},
  year={2023},
  publisher={Nature Publishing Group UK London}
}

@Article{Visentini-Scarzanella2017,
author={Visentini-Scarzanella, Marco
and Sugiura, Takamasa
and Kaneko, Toshimitsu
and Koto, Shinichiro},
title={Deep monocular {3D} reconstruction for assisted navigation in bronchoscopy},
journal={International Journal of Computer Assisted Radiology and Surgery},
year={2017},
month={Jul},
day={01},
volume={12},
number={7},
pages={1089-1099}}

@InProceedings{Rodriguez2022,
author="Rodriguez-Puigvert, Javier
and Recasens, David
and Civera, Javier
and Martinez-Cantin, Ruben",
title="{On the Uncertain Single-View Depths in Colonoscopies}",
booktitle="MICCAI",
year="2022",
publisher="Springer",
pages="130--140",
isbn="978-3-031-16437-8"
}

@article{LMS,
 ISSN = {01621459, 1537274X},
 author = {Peter J. Rousseeuw},
 journal = {Journal of the American Statistical Association},
 number = {388},
 pages = {871--880},
 publisher = {[American Statistical Association, Taylor & Francis, Ltd.]},
 title = {{Least Median of Squares Regression}},
 volume = {79},
 year = {1984} }

@InProceedings{Rodriguez-Puigvert_2023_ICCV,
    author    = {Rodr{\'\i}guez-Puigvert, Javier and Batlle, V{\'\i}ctor M. and Montiel, J.M.M. and Martinez-Cantin, Ruben and Fua, Pascal and Tard\'os, Juan D. and Civera, Javier},
    title     = {{LightDepth: Single-View Depth Self-Supervision from Illumination Decline}},
    booktitle = {(ICCV)},
    month     = {October},
    year      = {2023},
    pages     = {21273-21283}
}

@article{elvira2024,
  title={{CudaSIFT-SLAM}: multiple-map visual {SLAM} for full procedure mapping in real human endoscopy},
  author={Elvira, Richard and Tard{\'o}s, Juan D and Montiel, Jos{\'e} M M},
  journal={arXiv preprint arXiv:2405.16932},
  year={2024}
}

@article{open3d,
    author    = {Qian-Yi Zhou and Jaesik Park and Vladlen Koltun},
    title     = {{Open3D}: {A} Modern Library for {3D} Data Processing},
    journal   = {arXiv:1801.09847},
    year      = {2018},
}

@INPROCEEDINGS{liu2022sage,
  author={Liu, Xingtong and Li, Zhaoshuo and Ishii, Masaru and Hager, Gregory D. and Taylor, Russell H. and Unberath, Mathias},
  booktitle={2022 International Conference on Robotics and Automation (ICRA)}, 
  title={SAGE: SLAM with Appearance and Geometry Prior for Endoscopy}, 
  year={2022},
  volume={},
  number={},
  pages={5587-5593}}

@ARTICLE{nrslam,

  author={Rodríguez, Juan J. Gómez and Montiel, José M.M. and Tardós, Juan D.},

  journal={IEEE Transactions on Robotics}, 

  title={{NR-SLAM}: Nonrigid Monocular {SLAM}}, 

  year={2024},

  volume={40},

  number={},

  pages={4252-4264},
}

@article{wang2024endogslam,
        title={{EndoGSLAM: Real-Time Dense Reconstruction and Tracking in Endoscopic Surgeries using Gaussian Splatting}},
        author={Kailing Wang and Chen Yang and Yuehao Wang and Sikuang Li and Yan Wang and Qi Dou and Xiaokang Yang and Wei Shen},
        journal={arXiv preprint arXiv:2403.15124},
        year={2024}
    }

@ARTICLE{endodepth,

  author={Recasens, David and Lamarca, José and Fácil, José M. and Montiel, J. M. M. and Civera, Javier},

  journal={IEEE Robotics and Automation Letters}, 

  title={{Endo-Depth-and-Motion: Reconstruction and Tracking in Endoscopic Videos Using Depth Networks and Photometric Constraints}}, 

  year={2021},

  volume={6},

  number={4},

  pages={7225-7232},}

@inproceedings{paruchuri2024leveraging,
  title={{Leveraging Near-Field Lighting for Monocular Depth Estimation from Endoscopy Videos}},
  author={Paruchuri, Akshay and Ehrenstein, Samuel and Wang, Shuxian and Fried, Inbar and Pizer, Stephen M and Niethammer, Marc and Sengupta, Roni},
  booktitle={ECCV},
  year={2024},
  organization={Springer}
}

@ARTICLE{dso,

  author={Engel, Jakob and Koltun, Vladlen and Cremers, Daniel},

  journal={IEEE Transactions on Pattern Analysis and Machine Intelligence}, 

  title={Direct Sparse Odometry}, 

  year={2018},

  volume={40},

  number={3},

  pages={611-625},}

@inproceedings{wang2019recurrent,
  title={Recurrent neural network for (un-) supervised learning of monocular video visual odometry and depth},
  author={Wang, Rui and Pizer, Stephen M and Frahm, Jan-Michael},
  booktitle={CVPR},
  pages={5555--5564},
  year={2019}
}

@article{monodepth2,
  title     = {Digging into Self-Supervised Monocular Depth Prediction},
  author    = {Cl{\'{e}}ment Godard and
               Oisin {Mac Aodha} and
               Michael Firman and
               Gabriel J. Brostow},
  journal = {ICCV},
  month = {October},
year = {2019}
}

@article{Bonilla2024ARXIV,
  author    = {Sierra Bonilla and Shuai Zhang and Dimitrios Psychogyios and Danail Stoyanov and Francisco Vasconcelos and Sophia Bano},
  title     = {{Gaussian Pancakes: Geometrically-Regularized 3D Gaussian Splatting for Realistic Endoscopic Reconstruction}},
  journal   = {arXiv},
  year      = {2024},
}

@InProceedings{unet,
author="Ronneberger, Olaf
and Fischer, Philipp
and Brox, Thomas",
title="{U-Net: Convolutional Networks for Biomedical Image Segmentation}",
booktitle="MICCAI",
year="2015",
publisher="Springer",
pages="234--241",
isbn="978-3-319-24574-4"
}

@article{mahmoud2018live,
  title={Live tracking and dense reconstruction for handheld monocular endoscopy},
  author={Mahmoud, Nader and Collins, Toby and Hostettler, Alexandre and Soler, Luc and Doignon, Christophe and Montiel, Jose Maria Martinez},
  journal={IEEE transactions on medical imaging},
  volume={38},
  number={1},
  pages={79--89},
  year={2018},
  publisher={IEEE}
}

@article{murAcceptedTRO2015,
  title={{ORB-SLAM}: a Versatile and Accurate Monocular {SLAM} System},
  author={Mur-Artal, Ra\'ul AND Montiel, J. M. M. AND Tard\'os, Juan D.},
  journal={IEEE Trans. on Robotics},
  volume={31},
  number={5},
  pages={1147--1163},
  year={2015}
 }

@article{ORBSLAM3_TRO,
  title={{ORB-SLAM3}: An Accurate Open-Source Library for Visual, Visual-Inertial 
           and Multi-Map {SLAM}},
  author={Campos, Carlos AND Elvira, Richard AND Gomez, Juan J. AND Montiel, 
          Jos\'e M. M. AND Tard\'os, Juan D.},
  journal={IEEE Transactions on Robotics}, 
  volume={37},
  number={6},
  pages={1874-1890},
  year={2021}
 }

@inproceedings{wang2022neural,
  title     = {{Neural Rendering for Stereo 3D Reconstruction of Deformable Tissues in Robotic Surgery}},
  author    = {Wang, Yuehao and Long, Yonghao and Fan, Siu Hin and Dou, Qi},
  booktitle = {MICCAI},
  pages     = {431--441},
  year      = {2022},
  organization = {Springer}}

@inproceedings{batlle2023lightneus,
  title     = {{LightNeuS}: Neural Surface Reconstruction in Endoscopy Using Illumination Decline},
  author    = {Batlle, V{\'\i}ctor M. and Montiel, Jos{\'e} M. M. and Fua, Pascal and Tard{\'o}s, Juan D.},
  booktitle = {MICCAI},
  pages     = {503--513},
  year      = {2023},
  organization = {Springer},}

@inproceedings{Guo2024FreeSurGS,
  title={Free-SurGS: SfM-Free 3D Gaussian Splatting for Surgical Scene Reconstruction},
  author={Guo, Jiaxin and Wang, Jiangliu and Kang, Di and Dong, Wenzhen and Wang, Wenting and Liu, Yun-hui},
  booktitle={MICCAI},
  year={2024},
  organization={Springer}
}

@inproceedings{Sucar_2021_ICCV,
  author    = {Edgar Sucar and Shikun Liu and Joseph Ortiz and Andrew J. Davison},
  title     = {{iMAP}: {Implicit Mapping and Positioning in Real-Time}},
  booktitle = {ICCV},
  year      = {2021},
  pages     = {6229-6238}
}

@inproceedings{Wang2024StructurePreserving,
  author    = {Shuxian Wang and Akshay Paruchuri and Zhaoxi Zhang and Sarah McGill and Roni Sengupta},
  title     = {{Structure-preserving Image Translation for Depth Estimation in Colonoscopy Video}},
  booktitle = {MICCAI},
  year      = {2024},
  publisher = {Springer},
  volume    = {LNCS 15011},
  pages     = {667--677},
  month     = {October},
}

@article{luo2019details,
  title={Details preserved unsupervised depth estimation by fusing traditional stereo knowledge from laparoscopic images},
  author={Luo, Huoling and Hu, Qingmao and Jia, Fucang},
  journal={Healthcare Technology Letters},
  volume={6},
  number={6},
  pages={154},
  year={2019},
  publisher={Wiley-Blackwell}
}

@inproceedings{huang2021self,
  title={Self-supervised generative adversarial network for depth estimation in laparoscopic images},
  author={Huang, Baoru and Zheng, Jian-Qing and Nguyen, Anh and Tuch, David and Vyas, Kunal and Giannarou, Stamatia and Elson, Daniel S},
  booktitle={MICCAI},
  year={2021}
}

@inproceedings{NFL-BA,
  title={NFL-BA: Improving Endoscopic SLAM with Near-Field Light Bundle Adjustment},
  author={Dunn Beltran, Andrea and Rho, Daniel and Niethammer, Marc and Sengupta, Roni},
  booktitle={arXiv preprint arXiv:2412.13176},
  year={2024}
}

@inproceedings{morlana2024colonmapper,
  title={{ColonMapper}: topological mapping and localization for colonoscopy},
  author={Morlana, Javier and Tard{\'o}s, Juan D and Montiel, JMM},
  booktitle={IEEE ICRA},
  pages={6329--6336},
  year={2024}
}

@inproceedings{curless1996volumetric,
  title={A volumetric method for building complex models from range images},
  author={Curless, Brian and Levoy, Marc},
  booktitle={{Proceedings} of the 23rd annual conference on {Computer} graphics and interactive techniques},
  pages={303--312},
  year={1996}
}

@article{rau2019implicit,
title={Implicit domain adaptation with conditional generative adversarial networks for depth prediction in endoscopy},
author={Rau, Anita and Edwards, PJ Eddie and Ahmad, Omer F and Riordan, Paul and Janatka, Mirek and Lovat, Laurence B and Stoyanov, Danail},
journal={International Journal of Computer Assisted Radiology and Surgery},
pages={1--10},
publisher={Springer},
year={2019}
}

@inproceedings{karaoglu2021adversarial,
  title={Adversarial domain feature adaptation for bronchoscopic depth estimation},
  author={Karaoglu, Mert Asim and Brasch, Nikolas and Stollenga, Marijn and Wein, Wolfgang and Navab, Nassir and Tombari, Federico and Ladikos, Alexander},
  booktitle={MICCAI},
  pages={300--310},
  year={2021},
  organization={Springer}
}

@ARTICLE{olympusAR,

AUTHOR={Metzger, Rebecca  and Suppa, Per  and Li, Zhen  and Vemuri, Anant },

TITLE={Augmented reality navigation systems in endoscopy},

JOURNAL={Frontiers in Gastroenterology},

VOLUME={3},

YEAR={2024},

ISSN={2813-1169},

}

@article{luo2020consistent,
  title={Consistent video depth estimation},
  author={Luo, Xuan and Huang, Jia-Bin and Szeliski, Richard and Matzen, Kevin and Kopf, Johannes},
  journal={ACM Transactions on Graphics (ToG)},
  volume={39},
  number={4},
  pages={71--1},
  year={2020},
  publisher={ACM New York, NY, USA}
}

@inproceedings{izquierdo2023sfm,
  title={{SfM-TTR}: Using structure from motion for test-time refinement of single-view depth networks},
  author={Izquierdo, Sergio and Civera, Javier},
  booktitle={Proceedings of the IEEE/CVF Conference on Computer Vision and Pattern Recognition},
  pages={21466--21476},
  year={2023}
}

@misc{Lalaguna2024tfg,
  author      = {Alejandro Lalaguna},
  title       = {Medición computerizada del esófago de {Barrett}},
  type        = {Bachelor's thesis},
  institution = {University of Zaragoza},
  year        = {2024},
  url = {https://zaguan.unizar.es/record/149685}, 
}

\end{document}